\pdfoutput=1

\documentclass[11pt]{article}

\usepackage{acl}

\usepackage{times}
\usepackage{latexsym}

\usepackage[T1]{fontenc}

\usepackage[utf8]{inputenc}

\usepackage{microtype}

\usepackage{amsmath}
\usepackage{hyperref}
\usepackage{graphicx}
\usepackage{multirow}
\usepackage{CJKutf8}
\usepackage{ulem}
\usepackage{color}
\usepackage{booktabs}
\usepackage[symbol]{footmisc}
\usepackage{amssymb}
\usepackage{pifont}

\newcommand{\xmark}{\ding{55}}

\newcommand{\ja}[1]{\begin{CJK}{UTF8}{ipxm}#1\end{CJK}}

\newcommand{\ulblue}[1]{\textcolor{blue}{\underline{\color{black}{#1}}}}
\renewcommand{\thefootnote}{\fnsymbol{footnote}}

\title{Arukikata Travelogue Dataset with Geographic Entity Mention, Coreference, and Link Annotation}

\author{
Shohei Higashiyama${}^{1,2}$,
Hiroki Ouchi${}^{2,3}$,
Hiroki Teranishi${}^{3,2}$,
Hiroyuki Otomo${}^4$,\\
\textbf{Yusuke Ide${}^2$,~%
Aitaro Yamamoto${}^2$,~%
Hiroyuki Shindo${}^{2,3}$,~%
Yuki Matsuda${}^{2,3}$,}\\
\textbf{Shoko Wakamiya${}^{2}$,~%
Naoya Inoue${}^{5,3}$,~%
Ikuya Yamada${}^{6,3}$,~%
Taro Watanabe${}^2$}\\[3mm]
${}^1$NICT\quad%
${}^2$NAIST\quad%
${}^3$RIKEN\quad%
${}^4$CyberAgent, Inc.\quad%
${}^5$JAIST\quad%
${}^6$Studio Ousia\\[2mm]
\texttt{shohei.higashiyama@}${}^{*1}$\texttt{,}%
\texttt{\{hiroki.ouchi,ide.yusuke.ja6,}\\
\texttt{yamamoto.aitaro.xv6,shindo,yukimat,wakamiya,taro\}@}${}^{*2}$\texttt{,}\\
\texttt{hiroki.teranishi@}${}^{*3}$\texttt{,}%
\texttt{otomo\_hiroyuki@}${}^{*4}$\texttt{,}%
\texttt{naoya-i@}${}^{*5}$\texttt{,}%
\texttt{ikuya@}${}^{*6}$
}
\begin{document}
\maketitle
\begin{abstract}
Geoparsing is a fundamental technique for analyzing geo-entity information in text.
We focus on \textit{document-level} geoparsing, which considers geographic relatedness among geo-entity mentions, and presents a Japanese travelogue dataset designed for evaluating document-level geoparsing systems.
Our dataset comprises 200 travelogue documents with rich geo-entity information: 12,171 mentions, 6,339 coreference clusters, and 2,551 geo-entities linked to geo-database entries.
\end{abstract}
\setcounter{footnote}{1}
\footnotetext{
1:~\texttt{nict.go.jp},\quad
${}^{*}$2:~\texttt{is.naist.jp},\\
${}^{*}$3:~\texttt{riken.jp},\quad
${}^{*}$4:~\texttt{cyberagent.co.jp},\\
${}^{*}$5:~\texttt{jaist.ac.jp},\quad
${}^{*}$6:~\texttt{ousia.jp}}
\renewcommand*{\thefootnote}{\arabic{footnote}}

\section{Introduction}
\setcounter{footnote}{0}

Language expressions of locations or geographic entities (\textit{geo-entities}) are written in text to describe real-world events and human mobility. 
Thus, technologies for extracting and grounding geo-entity expressions are important for realizing various geographic applications.
For example, it is possible to recommend tourist spots and tour routes to users by analysis techniques for travelers' visited spots, taken routes, and reputation from text.

\textit{Geoparsing} ~\cite{leidner-2006-evaluation,gritta-etal-2020-pragmatic} is a fundamental technique that involves two subtasks: \textit{geotagging}, which identifies geo-entity mentions, and \textit{geocoding}, which identifies corresponding database entries for (or directly predicts the coordinates of) geo-entities.
Notably, geoparsing, geotagging, and geocoding can be regarded as special cases of entity linking (EL), named entity recognition (NER), and entity disambiguation (ED), respectively.

\begin{figure}[t]
\centering
\includegraphics{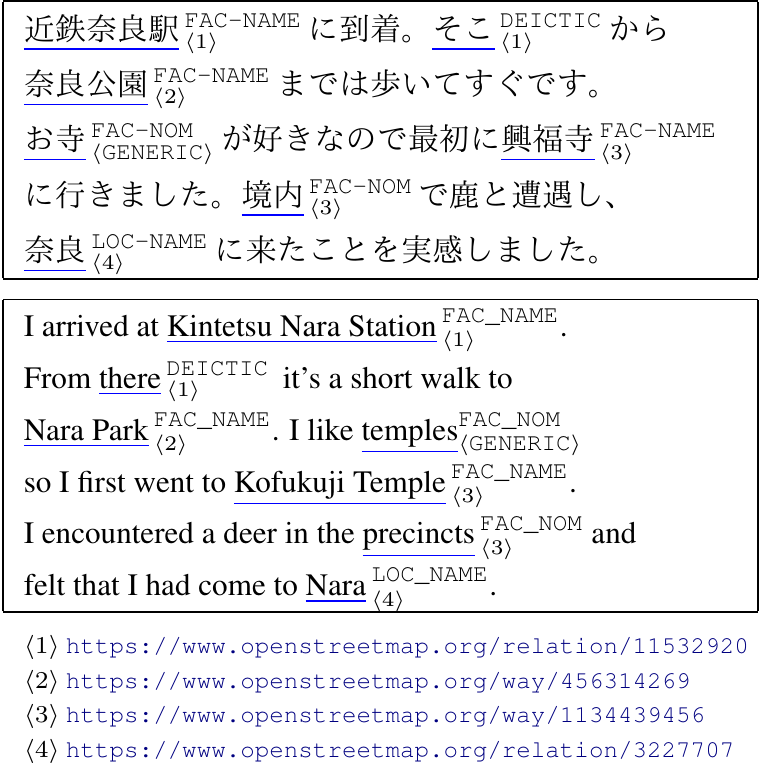}
\caption{Example illustration of an annotated travelogue document and its English translation. Expressions with blue underline indicate \ulblue{geo-entity mentions}, superscript strings (e.g., \texttt{FAC-NAME}) indicate entity types of mentions, subscript numbers (e.g., $\langle 1 \rangle$) indicate coreference cluster IDs of mentions. URLs outside the document indicate the corresponding entries in OpenStreetMap for coreference clusters.} \label{fig:anno_example}
\end{figure}

\begin{table*}[t]
  \centering
  \small{
  \begin{tabular}{llllcc} \toprule
  		  Dataset Name & Lang & Text Genre & Geo-database & Facility & Nominal \\ \midrule
      LGL Corpus~\cite{lieberman-etal-2010-geotagging} & en & News & GeoNames & \xmark & \xmark \\
      TR-News~\cite{kamalloo-etal-2018} & en & News & GeoNames & \xmark & \xmark \\
      GeoVirus~\cite{gritta-etal-2018-melbourne} & en & News & Wikipedia & \xmark & \xmark \\
      WikToR~\cite{gritta-etal-2018-whats} & en & Wikipedia & Wikipedia & \xmark & \xmark \\
      GeoCorpora~\cite{wallgrun-etal-2018-geocorpora} & en & Microblog & GeoNames & $\bigtriangleup$ & \xmark \\
      GeoWebNews~\cite{gritta-etal-2020-pragmatic} & en & News & GeoNames  & \xmark & $\checkmark$ \\
      LRE Corpus~\cite{matsuda-etal-2017-geographical} & ja & Microblog & CityBlocks \& Original & $\bigtriangleup$ & $\checkmark$  \\
      \midrule
      ATD-MCL (Ours) & ja & Travelogue & OpenStreetMap & $\checkmark$ & $\checkmark$ \\
                 \bottomrule
  \end{tabular}
  }
  \caption{Characteristics of representative geoparsing datasets and ours. The facility and nominal columns show the availability of geoparsed facility mentions and nominal mentions, respectively: $\checkmark$~(available), \xmark~(not available), and $\bigtriangleup$~(available to a limited extent). GeoCorpora has the very limited coverage of facility mentions. The gazetteer of facilities for the LRE Corpus has not been available due to licensing reasons. 
  }
  \label{tab:datasets}
\end{table*}

This study focuses on geoparsing from a perspective of \textit{document-level} analysis.
Geo-entity mentions that co-occur in a document tend to be geographically close to or related to each other; thus, information about some geo-entity mentions could help to specify information about other mentions.
For example, a mention \ja{興福寺} \textit{kofukuji} `Kofukuji Temple' shown in Figure~\ref{fig:anno_example} can be disambiguated to refer to the temple in Nara Prefecture, Japan, among the temples with the same name in different prefectures, by considering other mentions and the surrounding context that describe a trip to Nara.

This paper presents a dataset suitable for document-level geoparsing: Arukikata Travelogue Dataset with geographic entity Mention, Coreference, and Link annotation (ATD-MCL).
Specifically, we have designed the dataset to include three types of geo-entity information as illustrated in Figure~\ref{fig:anno_example}: (1) spans and entity types of geo-entity mentions, (2) coreference relations among mentions, and
(3) links from coreference clusters to corresponding entries in a geographic database.
To enable dataset users to perform (a) \textit{document-level} geoparsing, (b) evaluation of a \textit{broad coverage} of geo-entities, and (c) \textit{reproducible} experiments, we have set two design policies involving data source selection as below.

First, we adopt travelogues as a text genre with desirable characteristics for document-level geoparsing.
One characteristic is document length.
Travelogues can have enough lengths to include many geo-entity mentions and geographic relatedness among them, such as coreference and geographic proximity.
This is in contrast to short documents, e.g., tweets, used in some geoparsing datasets~\cite{matsuda-etal-2017-geographical,wallgrun-etal-2018-geocorpora}.
Another characteristic is geographic continuity among co-occurring mentions; mentions that refer to close real-world locations tend to appear in nearby positions within a document.
Since travel records reflect the actual travelers' movement trajectories to some extent, this characteristic is more notable in travelogues than other genres of documents, e.g., news articles used in representative geoparsing datasets~\cite{lieberman-etal-2010-geotagging,kamalloo-etal-2018,gritta-etal-2018-melbourne,gritta-etal-2020-pragmatic}.
Based on the above background, we use the Arukikata Travelogue Dataset (ATD)\footnote{\url{https://www.nii.ac.jp/dsc/idr/arukikata/}}~\cite{arukikata-2022,ouchi-etal-2023-atd}, which was constructed from user-posted travelogues in Japanese and is provided to research institutions for research purposes.

Second, we aim to cover diverse variations of geo-entity mentions.
From a semantic perspective, we target not only coarse-grained locations, such as countries and municipalities, but also fine-grained locations and facilities, including districts, amenity buildings, landmarks, roads, and public transport lines.
From a syntactic perspective, we target not only named entities (NEs) written with proper noun phrases (NPs), but also nominal and demonstrative expressions that can refer to real-world locations.
So far, as summarized in Table~\ref{tab:datasets},
it has been a challenge to achieve a high coverage particularly for facility entity mentions mainly because of the limited coverage of public geo-databases, e.g., GeoNames\footnote{\url{https://www.geonames.org/}}.
To address this database coverage problem, we adopt OpenStreetMap\footnote{\url{https://www.openstreetmap.org/}} (OSM), a free, editable, and large-scale geo-database of the world.
The usefulness of OSM has been steadily increasing every year, as evidenced by the growth in the number of registered users, which rose from 1M in 2013 to 10M in 2023, and the increase in node entries,\footnote{Node is the most fundamental element that consists of a single point in space among OSM entry types.} which soared from over 1.5B in 2013 to over 80B in 2023.\footnote{\url{https://wiki.openstreetmap.org/wiki/Stats}}
Our evaluation have actually demonstrated that OSM had the reasonable coverage for our dataset; 74\% of coreference clusters that contained facility NE mentions were able to link to OSM entries.


Selecting these data sources, namely, ATD and OSM, also fulfills our third intention, i.e., ensuring the reproducible experiments for other researchers, through the public release of our annotated data.\footnote{We will release our dataset at \url{https://github.com/naist-nlp/atd-mcl}.}

As a result of manual annotation work, our dataset comprises 12,273 sentences, from the full text of 200 travelogue documents, with rich geo-entity information: 12,171 geo-entity mentions, 6,339 coreference clusters (geo-entities), and 2,551 linked geo-entities.\footnote{We conducted link annotation for 100 out of 200 documents including 3,208 geo-entities as described in \S\ref{sec:data_stat}.}
Furthermore, our investigation of inter-annotator agreement (IAA) for mention, coreference, and link annotation suggest the practical quality of our dataset in terms of the consistency.

\section{Overview of Annotation Process}
The authors conducted trial annotation of a small number of documents and defined annotation guidelines for three types of information: mention, coreference, and link.
We then asked annotators in a data annotation company to perform the annotation work.\footnote{Five or six annotators along with an annotation manager performed the actual annotation work, in each step of mention, coreference, and link annotation.}
Data preparation by the authors and annotation work with three steps by the annotators were performed as described below.

\paragraph{Data Preparation}
We first picked up the documents about domestic travel within reasonable document length, 4,000 characters, from ATD.
Then, we applied NFKC normalization except for whitespace characters that were converted to full-width.
In addition, we applied the GiNZA NLP Library\footnote{\url{https://github.com/megagonlabs/ginza}}~\cite{ginza-2019} to raw sentences for sentence segmentation and automatic annotation of named entity (NE) mention candidates.

\paragraph{Mention Annotation Step}
For input documents, annotators identify spans of geo-entity mentions, which can refer to real-world locations, and assign predefined entity type tags to identified mentions by modifying the automatic annotation results.
We adopted brat\footnote{\url{https://github.com/nlplab/brat}}~\cite{stenetorp-etal-2012-brat} as the annotation tool for mention and coreference annotation.

\paragraph{Coreference Annotation Step}
Annotators identify the groups of geo-entity mentions that corefer to the same locations for each document. 
Using brat's function of relation annotation between two mentions, annotators need to assign relation edges to mention pairs that should be included in the same coreference cluster.

\paragraph{Link Annotation Step}
Annotators link each coreference cluster to the URL of the corresponding OSM entry (for example, $\langle 1 \rangle$--$\langle 4 \rangle$ in Figure~\ref{fig:anno_example}) on the basis of OSM and web search results.
For this step, we converted brat output files with mention and coreference information to TSV files, where each row represents a coreference cluster or a member mention.
Annotators fill URLs in the specific cells in the TSV files.

\section{Annotation Guidelines}

\subsection{Mention Annotation}
In the mention annotation step, the \textit{entity types} and \textit{spans} of geo-entity mentions are identified. 

\begin{table}[t]
\centering
\small
\begin{tabular}{l|l}
\toprule
Type and subtype & Example mentions\\
\midrule
\texttt{LOC-NAME} & \ja{奈良} `Nara'; \ja{生駒山} `Mt. Ikoma'\\
\texttt{LOC-NOM}  & \ja{町} `town'; \ja{島} `island'\\
\midrule
\texttt{FAC-NAME} & \ja{大神神社} `\=Omiwa Shrine'\\
\texttt{FAC-NOM}  & \ja{駅} `station'; \ja{公園} `park'\\
\midrule
\texttt{TRANS-NAME} & \ja{特急ひのとり} `Ltd. Exp. Hinotori'\\
\texttt{TRANS-NOM}  & \ja{バス} `bus'; \ja{フェリー} `ferry'\\
\midrule
\texttt{LINE-NAME} & \ja{近鉄奈良線} `Kintetsu Nara Line'\\
\texttt{LINE-NOM}  & \ja{国道} `national route'; \ja{川} `river'\\
\bottomrule
\end{tabular}
\caption{Example mentions of main entity types.}
\label{tab:etype_example}
\end{table}

\paragraph{Entity Type}
We target geo-entity types corresponding to \texttt{Location}, \texttt{Facility}, and \texttt{Vehicle} in Sekine's Extended Named Entity (ENE) hierarchy~\cite{sekine-etal-2002-extended} version 9.0,\footnote{\url{http://ene-project.info/ene9/?lang=en}} but exclude \texttt{Astronomical\_Object}.
For our dataset, we define eight entity types: \texttt{LOC}, \texttt{FAC}, \texttt{TRANS}, \texttt{LINE}, \texttt{LOC\_ORG}, \texttt{FAC\_ORG}, \texttt{LOC\_OR\_FAC}, \texttt{DEICTIC}.
Specifically, \texttt{LOC}, \texttt{FAC}, and \texttt{TRANS} represent location, facility, and public transport vehicle. 
\texttt{LINE} represents road, waterway/river,\footnote{We regard river names as a case of \texttt{LINE}, whereas the ENE defines \texttt{River} as a subtype of \texttt{Location}.} or public transport line.
These four types are further divided into two subtypes, i.e., \texttt{NAME} and \texttt{NOM}, corresponding to whether a mention is named or nominal, as shown in Table~\ref{tab:etype_example}.
\texttt{LOC\_ORG} and \texttt{FAC\_ORG} indicate location and facility mentions, respectively, that metonymically refer to organizations, e.g., \ulblue{\ja{ホテル}} \textit{hoteru} in a sentence such as ``The \ulblue{hotel} serves its lunch menu.''
\texttt{LOC\_OR\_FAC} indicates nominal mentions that can refer to both location and facility, e.g., \ulblue{\ja{観光地}} \textit{kank\=ochi} in a sentence such as ``I like this \ulblue{sightseeing spot}.''
Lastly, \texttt{DEICTIC} indicates deictic expressions that refer to other geo-entity mentions or geo-entities in the real world, 
e.g., \ulblue{\ja{そこ}} \textit{soko} in a sentence such as ``I walked for 10 minutes from \ulblue{there}.''

\begin{table}[t]
\renewcommand{\arraystretch}{1.2}
\centering
\small
\begin{tabular}{ll}
\toprule
\multirow{2}{*}{(a)} 
& \ulblue{[\ja{山頂}]${}_{m}$ [\ja{駐車場}]${}_{h}$} \\
& \ulblue{[parking area]${}_{h}$ [on top of the mountain]${}_{m}$}\\
\midrule
\multirow{2}{*}{(b)} 
& \ulblue{\ja{[駅ビル]${}_{n_1}$ [「ビエラ奈良」]}${}_{n_2}$} \\
& \ulblue{[station building]${}_{n_1}$ [Vierra Nara]${}_{n_2}$}\\
\midrule
\multirow{2}{*}{(c)} 
& \ulblue{\ja{天国への階段}}\\
& \ulblue{Stairway to Heaven}\\
\midrule
\multirow{2}{*}{(d-1)} 
& \ulblue{[\ja{東}]${}_{a}$ [\ja{東京}]}\\
& \ulblue{[East]${}_{a}$ [Tokyo]}\\
\midrule
\multirow{2}{*}{(d-2)} 
& {\ulblue{[\ja{北海道}]} [\ja{全域}]${}_{a}$}\\
& {[the whole area of]${}_{a}$ \ulblue{[Hokkaido]}}\\
\midrule
\multirow{2}{*}{(e-1)} 
& \ulblue{[\ja{京都}]${}_{m}$} [\ja{旅行}]${}_{h}$\\
& \ulblue{[Kyoto]${}_{m}$} [Travel]${}_{h}$\\
\midrule
\multirow{2}{*}{(e-2)} 
& {[\ja{三輪}]${}_{m}$ [\ja{そうめん}]${}_{h}$}\\
& {[Miwa]${}_{m}$ [somen noodles]${}_{h}$}\\
\midrule
\multirow{2}{*}{(f)} 
& [\ulblue{[\ja{保津川}]${}_{g}$} \ja{下り}]${}_{n}$\\
& [\ulblue{[Hozugawa river]${}_{g}$} boat tour]${}_{n}$\\
\bottomrule
\end{tabular}
\renewcommand{\arraystretch}{1.0}
\caption{Examples of \ulblue{mention spans}.}
\label{tab:span_example}
\end{table}

\paragraph{Mention Span} The spans of mentions are determined as follows.
Generally, an NP where a head $h$ is modified by a nominal modifier $m$ is treated as a single mention (Table~\ref{tab:span_example}-a).
An appositive compound of two nouns $n_1$ and $n_2$ is treated as a single mention (\ref{tab:span_example}-b) unless some expressions (e.g., \textit{no}-particle ``\ja{の}'') or separator symbols (e.g., \textit{t\=oten} ``\ja{、}'') are inserted between them. 
A common name is treated as a single mention even if it is not a simple NP (\ref{tab:span_example}-c).
For an NP with an affix or affix-like noun $a$ representing directions or relative positions, a cardinal direction prefix preceding a location name is included in the span (\ref{tab:span_example}-d-1) but other affixes are excluded from the span (\ref{tab:span_example}-d-2).
In the case that a modifier $m$ represents a geo-entity but its NP head $h$ does not, the modifier is treated as a single mention if the head is a verbal noun that means move, stay, or habitation (\ref{tab:span_example}-e-1),
but the NP is not treated as a mention if not (\ref{tab:span_example}-e-2).
In the case that a geo-entity name $g$ is embedded in a non-geo-entity mention $n$, the inner geo-entity name is treated as a geo-entity mention if the external entity corresponds to an event held in the real world (\ref{tab:span_example}-f), but it is not treated as a geo-entity mention if the external entity corresponds to other types of entities, such as an organization and a title of a work.

\subsection{Coreference Annotation}
The coreference annotation step requires to assign mention-level \textit{specificity tags} or mention-pair-level \textit{relations} to mentions identified in the previous step except for those labeled with \texttt{TRANS} tags.\footnote{We set coreference and link annotation for \texttt{TRANS} mentions as outside the scope of this study because it is not obvious how to treat the identity of those mentions and OSM does not contain such type of entries.
However, \texttt{TRANS} (\texttt{-NAME}) mentions may be helpful to identify the referents of other types of mentions that are not clearly written.}
Specificity tags include \texttt{GENERIC} and \texttt{SPEC\_AMB}.
\texttt{GENERIC} is assigned to a generic mention, e.g., \ja{お寺} \textit{otera} `temples' in Figure~\ref{fig:anno_example}, to distinguish singleton mentions, which refer to real-world location but do not corefer with other mentions.
\texttt{SPEC\_AMB} is assigned to a mention that refers to a specific real-world location but is ambiguous as to which detailed area it refers to, e.g., \ulblue{\ja{海}} \textit{umi} in a sentence such as ``You can see a beautifull \ulblue{sea} from this spot.''
After (or concurrently with) specificity tag annotation, a relation, which is either \texttt{COREF} or \texttt{COREF\_ATTR}, is assigned to a pair of mentions holding the relation both of which have been labeled with neither specificity tag.

\paragraph{Identical Coreference}
A coreference relation \texttt{COREF} is assigned to two mentions both of which refer to the same real-wold location, e.g., \ja{近鉄奈良駅} \textit{kintetsu nara eki} `Kintetsu Nara Station' and \ja{そこ} \textit{soko} `there' in Figure~\ref{fig:anno_example} $\langle 1 \rangle$.
After relation annotation, a set of mentions sequentially connected through binary \texttt{COREF} (or \texttt{COREF\_ATTR}) relations is regarded as a single coreference cluster.
A mention without any relation or specificity tag is regarded as a singleton, e.g., mentions in Figure~\ref{fig:anno_example} $\langle 2 \rangle$ and $\langle 4 \rangle$.\footnote{Although we also mark singleton mentions with coreference cluster IDs in Figure~\ref{fig:anno_example} for clarity, singletons were not annotated with any coreference information in the actual work.} 

\paragraph{Attributive Coreference}
A directed relation \texttt{COREF\_ATTR} is assigned to mentions, either of which expresses the attribute of the other, in appositive phrases or copular sentences.
For example, a sentence in Figure~\ref{fig:coref_attr} is annotated with \texttt{COREF\_ATTR} relations from mention 2 to mention 1 and from mention 2 to mention 3.
This schema is similar to that in WikiCoref~\cite{ghaddar-langlais-2016-wikicoref} and enables to distinguish attributive coreference from identical coreference.

\begin{figure}[t]
\centering
\small
\renewcommand{\arraystretch}{1.5}
\begin{tabular}{|l|}\hline
\ja{${}^{1}$\ulblue{世界遺産}・${}^{2}$\ulblue{白川郷}は素敵な${}^{3}$\ulblue{ところ}でした。}\\
A ${}^{1}$\ulblue{world heritage site}, ${}^{2}$\ulblue{Shirakawago} was a nice ${}^{3}$\ulblue{place}.\\
\hline
\end{tabular}
\renewcommand{\arraystretch}{1.0}
\caption{Examples of \ulblue{attributive mentions}.} \label{fig:coref_attr}
\end{figure}

\paragraph{Other Cases}
For mentions whose referents are geographically overlapped but not identical, any coreference relations are not assigned.
For example, \ulblue{\ja{首都高速道路}} \textit{shuto k\=osoku d\=oro} `Metropolitan Expressway' and \ulblue{\ja{湾岸線}} \textit{wangansen} `Bayshore Route,' which have a whole-part relation, are not regarded as coreferring mentions.


\subsection{Link Annotation} \label{sec:guideline_link}
Link annotation is done in the following process.
(1)  A unique number indicating a coreference ID is automatically assigned to each coreference cluster (including singleton) in a document.
(2) For each coreference cluster, an annotator determines one or more normalized names \texttt{NORM\_NAMES} of the referent location, e.g., formal or common name.
This can be simply done by selecting a proper name mention string among cluster member mentions in many cases.
(3) The annotator searches and assigns an appropriate OSM entry URL to the coreference cluster using search engines.\footnote{Because it was sometimes difficult to find desired entries by the search engine in the OSM's official site, we asked the annotators to also use additional search engines: Google search, Wikidata search, and our original search engine that we developed.}
The specific assignment process of entries is as follows:
\begin{itemize}
\setlength{\parskip}{0cm}
\setlength{\itemsep}{0cm}
\item If one or more candidate entries for a coreference cluster are found, assign the most probable candidate as \texttt{BEST\_URL} and (up to two) other possible candidates as \texttt{OTHER\_URLS}.
\item If only a candidate entry that geographically includes but does not exactly match with the real-world referent is found, assign the found entry with \texttt{PART\_OF} tag.
\item If no candidate entries are found in OSM, search and assign an appropriate entry in alternative databases: Wikidata,\footnote{\url{https://www.wikidata.org/}} Wikipedia,\footnote{\url{https://ja.wikipedia.org/}} and general web pages describing the real-world referent.\footnote{These auxiliary databases enable to preserve referent information for cases where expected entries are not present in OSM (as of the time of annotation).}
\item If no candidate entries are found in any databases, assign \texttt{NOT\_FOUND} tag instead of an entry URL.
\end{itemize}

In addition, we set the following general policies in this step:
\begin{itemize}
\item As context to identify the entry for a coreference cluster, annotators are required to consider paragraphs where member mentions occur but allowed to consider broader context.
\item Annotators are allowed to merge or split clusters, which have been annotated in the coreference annotation step, on the basis of their interpretation.
\item Annotators can skip the searching steps and assign \texttt{NOT\_FOUND} tag to a coreference cluster when all member mentions and surrounding context have no specific information to identify the referent.
\end{itemize}

% ...

\section{Dataset Statistics} \label{sec:data_stat}

\begin{table}[t]
\small
\centering
\begin{tabular}{l|rrrrr}
\toprule
      & \multicolumn{1}{c}{\#Doc} & \multicolumn{1}{c}{\#Sent} & \multicolumn{1}{c}{\#Char} & \multicolumn{1}{c}{\#Men} & \multicolumn{1}{c}{\#Ent} \\
\midrule
Set-A &  100 & 5,949 & 139,406 & 6,052 & 3,131\\
Set-B &  100 & 6,324 & 141,548 & 6,119 & 3,208\\
\midrule
Total & 200 & 12,273 & 280,954 & 12,171 & 6,339\\
\bottomrule
\end{tabular}
\caption{Statistics of the ATD-MCL.} \label{tab:data_stat}
\end{table}

\subsection{Basic Statistics}
Annotators first annotated 200 documents with mention information, then annotated the same 200 documents with coreference information, and finally annotated 100 of those documents with link information.
We call the latter 100 documents with link annotation as Set-B and the remaining 100 documents without link annotation as Set-A.
We show the numbers of documents (\#Doc), sentences (\#Sent), characters (\#Char), mentions (\#Men), and entities (coreference clusters) (\#Ent) in the ATD-MCL in Table~\ref{tab:data_stat}.

\begin{table}[t]
\small
\centering
\begin{tabular}{l|rrrrr}
\toprule
      & \multicolumn{1}{c}{\texttt{LOC}} & \multicolumn{1}{c}{\texttt{FAC}} & \multicolumn{1}{c}{\texttt{LINE}} & \multicolumn{1}{c}{\texttt{TRANS}} & \multicolumn{1}{c}{GeoOther} \\
\midrule
\texttt{NAME}  & 2,289 & 3,239 & 462 & 257 & --\\
\texttt{NOM}   & 861 & 2,851 & 582 & 666 & --\\
Other & -- & -- & -- & -- & 907\\
\midrule
Total & 3,150 & 6,090 & 1,044 & 923 & 907\\
\bottomrule
\end{tabular}
\caption{Tag distribution of geo-entity mentions in the whole dataset. ``GeoOther'' mentions consist of 372 \texttt{LOC\_OR\_FAC} and 535 \texttt{DEICTIC} mentions.} \label{tab:dist_geo_men}
\end{table}

\begin{table}[t]
\footnotesize
\centering
\begin{tabular}{l|rrrrrrr}
\toprule
Size & 1 & 2 & 3 & 4 & 5 & 6 & $\geq 7$\\
\midrule
\#Cls  & 4,083 & 1,278 & 507 & 240 & 103 & 58 & 70 \\
\#Typ & 1.0 & 1.5 & 2.0 & 2.3 & 2.6 & 2.8 & 3.3\\
\bottomrule
\end{tabular}
\caption{Number of geo-entity coreference clusters (\#Cls) and the average number of member mention text types (\#Typ) for each size.} \label{tab:dist_ent_size}
\end{table}

\begin{table}[t]
\small
\centering
\begin{tabular}{l|rrrrr}
\toprule
      & \multicolumn{1}{c}{{\texttt{LOC}}} & \multicolumn{1}{c}{\texttt{FAC}} & \texttt{LINE} & \texttt{MIX} & \texttt{UNK} \\
\midrule
Set-A & 819 & 1,823 & 327 & 29 & 133 \\
Set-B & 852 & 1,819 & 370 & 22 & 145\\
\midrule
Total & 1,671 & 3,642 & 697 & 51 & 278 \\
\bottomrule
\end{tabular}
\caption{Tag distribution of geo-entities.} \label{tab:dist_geo_ent}
\end{table}

\subsection{Mention Annotation}
In the mention annotation step, 12,171 mentions were identified; they consist of 12,114 geo-entity and 57 non-geo-entity mentions (23 \texttt{LOC\_ORG} and 34 \texttt{FAC\_ORG} mentions).
Table~\ref{tab:dist_geo_men} shows the distribution of geo-entity mentions for entity type tags.
The tag distribution represents some characteristics of travelogue documents of our dataset.
First, the documents contains the largest number of facility mentions, which is even more than the number of location mentions.
Second, the documents also contains the similar number of non-\texttt{NAME} (5,867)\footnote{Non-\texttt{NAME} mentions include \texttt{LOC\_OR\_FAC}, and \texttt{DEICTIC} mentions, in addition to all \texttt{NOM} mentions.} to \texttt{NAME} mentions (6,247).

\subsection{Coreference Annotation}
As a result of the coreference annotation step, 289 \texttt{GENERIC} mentions and 322 \texttt{SPEC\_AMB} mentions along with 923 \texttt{TRANS} mentions were excluded from the coreference relation annotation. 
Out of the remaining 10,580 
mentions, 6,497 mentions were annotated with one or more \texttt{COREF} and/or \texttt{COREF\_ATTR} relations among other mentions, of which 350 mention pairs were annotated with \texttt{COREF\_ATTR} relations.
These mentions comprise coreference clusters with size $\geq\!2$, and the remaining 4,083 mentions correspond to singletons.
Table~\ref{tab:dist_ent_size} shows the number of clusters and the average number of mention text types (distinct strings) among members\footnote{For example, for clusters $C_1=\{$``Nara Station'', ``Nara Sta.'', ``Nara''$\}$ and $C_2=$\{``Kyoto Pref.'', ``Kyoto'', ``Kyoto''$\}$, the numbers of distinct member mention strings are three and two, respectively, and their average is 2.5.} for each cluster size.
This indicates that 35.6\% (2,256/6,339) of coreference clusters have more than one member; that is, multiple mentions in a document often refer to the same referent.

In addition, we automatically assign an entity type tag to each coreference cluster, i.e., entity, from the tags of its member mentions.\footnote{(a) \texttt{LOC}, \texttt{FAC}, or \texttt{LINE} is assigned to an entity that the members' tags include only one of the three types and optionally include \texttt{LOC\_OR\_FAC} or \texttt{DEICTIC}. (b) \texttt{UNK} is assigned to an entity that all members' tags are \texttt{LOC\_OR\_FAC} or \texttt{DEICTIC}. 
(c) \texttt{MIX} is assigned to an entity that the members' tags include two or three of \texttt{LOC}, \texttt{FAC}, and \texttt{LINE}.}
Table~\ref{tab:dist_geo_ent} shows the tag distribution of entities, which is similar to the tag distribution of mentions shown in Table~\ref{tab:dist_geo_men}.

\begin{table}[t]
\small
\centering
\begin{tabular}{l|r|rr}
\toprule
      & \multicolumn{1}{c|}{All} & HasRef & HasOSMRef \\
\midrule
HasName     & 2,001 & 1,942 & 1,574 \\
HasNoName & 1,207 & 609   & 485   \\
\midrule
Total         & 3,208 & 2,551 & 2,059 \\
\bottomrule
\end{tabular}
\caption{Numbers of Set-B entities that have names and/or references in the \texttt{PART\_OF}-inclusive setting where entities assigned with \texttt{PART\_OF} (along with URLs) are counted as instances of ``Has(OSM)Ref.''} \label{tab:dist_link}
\end{table}

\begin{table}[t]
\small
\centering
\begin{tabular}{l|r|rr}
\toprule
      & \multicolumn{1}{c|}{All} & HasRef & HasOSMRef \\
\midrule
HasName     & 2,001 & 1,861 & 1,514 \\
HasNoName & 1,207 & 298   & 221   \\
\midrule
Total         & 3,208 & 2,159 & 1,735 \\
\bottomrule
\end{tabular}
\caption{Numbers of Set-B entities that have names and/or referents in the \texttt{PART\_OF}-exclusive setting where entities assigned with \texttt{PART\_OF} (along with URLs) are NOT counted as instances of ``Has(OSM)Ref.''} \label{tab:dist_link_nopartof}
\end{table}

\subsection{Link Annotation}
As shown in Table~\ref{tab:dist_link}, in the link annotation step for Set-B, 79.5\% (2,551) and 64.2\% (2,059) of 3,208 entities have been annotated with any URLs and OSM entry URLs, respectively, including entities annotated with \texttt{PART\_OF} tags.
For ``HasName'' entities in which at least one member mention is labeled as \texttt{NAME}, any and OSM entry URLs are assigned to 97.1\% (1,942/2,001) and 78.7\% (1,574/2,001) of them, respectively. 
This indicates that the real-world referents can be easily identified for most of the entities explicitly written with their names.
For the remaining ``HasNoName'' entities, no any and OSM entry URLs are assigned to 50.5\% (609/1,207) and 40.2\% (485/1,207) of them, respectively.
This suggests that identifying the referents from unclearly written mentions and context is difficult even for humans.

As shown in Table~\ref{tab:dist_link_nopartof}, the percentages of referent-identified entities decrease in the setting where entities assigned with \texttt{PART\_OF} are excluded.
The result indicates the reasonable coverage of OSM for various types of locations in Japan.
Overall, entities assigned with OSM entries account for 75.7\% (1,514/2,001) of ``HasName'' entities.
For details on each entity type tag of \texttt{LOC}, \texttt{FAC}, \texttt{LINE}, and the others, entities assigned with OSM entries account for 79.3\% (811/1,096), 74.0\% (544/686), 72.7\% (144/198), and 71.4\% (15/21) of ``HasName'' entities with the specified tag, respectively.

\subsection{Summary}
Our analysis showed the statistical characteristics of our dataset as follows.
(1) Facility mentions account for 50.3\% and nominal or demonstrative expressions account for 48.4\% of geo-entity mentions, respectively.
(2) Multi-member clusters account for 35.6\% of coreference clusters, suggesting that the same geo-entity is repeatedly referred to by different expressions in a document.
(3) Geo-entities assigned with OSM entries account for 75.7\% of entities with NE mentions (\texttt{PART\_OF}-exclusive setting), indicating the reasonable coverage of OSM for various types of locations in Japan.

\section{Inter-Annotator Agreement Evaluation} \label{sec:iaa}

For mention, coreference, and link annotation, we requested two annotators to independently annotate the same 10, 10, and 5 documents out of 200, 200, and 100 documents, respectively. 
We measured IAA for the three annotation tasks.

\subsection{Mention Annotation}

\begin{table}[t]
\small
\centering
\begin{tabular}{l|rrrr|rr}
\toprule
\multirow{2}{*}{Tag set} & \multicolumn{4}{c|}{Token} & \multicolumn{2}{c}{Type} \\
& \multicolumn{1}{c}{F1} & \multicolumn{1}{c}{\#W1} & \multicolumn{1}{c}{\#W2} & \multicolumn{1}{c|}{\#M} & \multicolumn{1}{c}{\#W1} & \multicolumn{1}{c}{\#W2} \\
\midrule
*-\texttt{NAME} & 0.835 & 229 & 243 & 197 & 162 & 174 \\
*-\texttt{NOM} & 0.867 & 195 & 197 & 170& 97 & 106 \\
\texttt{L\_O\_F} & 0.552 & 19 & 10 & 8 & 8 & 5 \\
\texttt{DEICT} & 0.621 & 19 & 10 & 9 & 6 & 3 \\
\texttt{L\_ORG} & -- & 0 & 0 & 0 & 0 & 0 \\
\texttt{F\_ORG} & 0 & 1 & 0 & 0 & 1 & 0 \\
\midrule
All & 0.832 & 463 & 460 & 384& 274 & 283 \\
\bottomrule
\end{tabular}
\caption{Inter-annotator agreement for mention annotation. *-\texttt{NAME}, *-\texttt{NOM}, \texttt{L\_O\_F}, \texttt{DEICT}, \texttt{L\_ORG}, and \texttt{F\_ORG} indicate all \texttt{NAME} mentions, all \texttt{NOM} mentions, \texttt{LOC\_OR\_FAC}, \texttt{DEICTIC}, \texttt{LOC\_ORG}, and \texttt{FAC\_ORG}, respectively. The token and type columns indicate the scores and numbers based on token and type frequencies of mention text, respectively. ``M'' indicates matched mention tokens between two annotators.} \label{tab:iaa_mention}
\end{table}

As an IAA measure for mention annotation, we calculated F1 scores between results of two annotators (W1 and W2), based on exact match of both spans and tags.\footnote{The F1 scores in this setting were the same as the F1 scores in the setting of only exact span match; there were no mentions with matched span and mismatched tags between the annotators.}
Table~\ref{tab:iaa_mention} shows the F1 score for each tag set and the numbers of annotated mentions by W1, W2, and both (M).

We obtained F1 score of 0.832 for all mentions. 
Higher F1 score of 0.867 for \texttt{NOM} mentions than that of 0.835 for \texttt{NAME} mentions is probably because less variety of \texttt{NOM} mention text types eased the annotation work for those mentions, as suggested by the mention token/type frequencies in Table~\ref{tab:iaa_mention}.

\begin{table}[t]
\small
\centering
\begin{tabular}{l|c|cccc}
\toprule
$|C|$ & \#W1/\#W2 & \multicolumn{1}{c}{MUC} & \multicolumn{1}{c}{B${}^3$} & \multicolumn{1}{c}{CEAF${}_e$} & \multicolumn{1}{c}{Avg.} \\
\midrule
\multicolumn{6}{c}{(a) Original clusters with all mentions} \\
\midrule
$\geq\!1$ & 237/297 & 0.913 & 0.878 & 0.782 & 0.858\\
$\geq\!2$ &  91/79 & 0.797 & 0.768 & 0.811 & 0.792\\
\midrule
\multicolumn{6}{c}{(b) Clusters only with \texttt{NAME} mentions} \\
\midrule
$\geq\!1$ & 237/297 & 0.959 & 0.935 & 0.893 & 0.929\\
$\geq\!2$ &  91/79 & 0.912 & 0.868 & 0.844 & 0.874\\
\bottomrule
\end{tabular}
\caption{Inter-annotator agreement between the two annotators for coreference clusters in coreference annotation. 
The top two rows and the bottom two rows indicate (a) the results for the original coreference clusters and (b) the results for the clusters where only \texttt{NAME} mentions are retained, respectively.
(i) $|C|\geq\!1$ and (ii) $|C|\geq\!2$ indicate clusters $C$ with the specified size; the former includes singletons but the latter does not.
The scores in the columns of MUC, B${}^3$, and CEAF${}_e$ are F1 scores for each metrics.
}
\label{tab:iaa_coref_cls}
\end{table}

\subsection{Coreference Annotation}
Ten documents annotated by two annotators did not include any mentions with \texttt{GENERIC} tag, \texttt{SPEC\_AMB} tag, or mention pairs with \texttt{COREF\_ATTR} relation.

As IAA measures for \texttt{COREF} relation annotation, we used metrics commonly used in coreference resolution studies: MUC \cite{vilain-etal-1995-model}, B${}^3$ \cite{bagga-etal-1998-algorithms}, CEAF${}_e$ \cite{luo-2005-coreference}, and the average of the three metrics (a.k.a CoNLL score) \cite{pradhan-etal-2012-conll}.
Table~\ref{tab:iaa_coref_cls} shows F1 scores between two annotators' (W1 and W2) results for each IAA measure and the numbers of clusters constructed from two annotators' results for 2$\times$2 settings: consider (a) original coreference clusters with all mentions or (b) clusters where only \texttt{NAME} mentions are retained (i.e., whether non-\texttt{NAME} mentions are included or not), and (i) clusters with size $\geq 1$ or (ii) clusters with size $\geq 2$ (i.e., whether singletons are included or not).

In the basic setting (a)-(i), we obtained the average F1 score of 0.858.
In addition, we observed two intuitive results. 
One is the lower scores for (a) than for (b), indicating that it was difficult to identify which mentions coreferred with non-\texttt{NAME} mentions.
The other is the higher scores for (i) than for (ii); this is because leaving mentions as singletons is more likely to agree, since each mention is a singleton by default.

\subsection{Link Annotation}

\begin{table}[t]
\small
\centering
\begin{tabular}{r|c|cc|cc}
\toprule
\multirow{2}{*}{In/OO-DB} & \multirow{2}{*}{\#W1/\#W2} & \multicolumn{2}{c|}{(a) Original} & \multicolumn{2}{c}{(b) Grouped} \\
& & \#M & F1 & \#M & F1 \\
\midrule
In-OSM & 70/63 & 50 & 0.752 & 56 & 0.842\\
OO-OSM & 27/34 & 26 & 0.852 & 26 & 0.852\\
\midrule
In-Any & 81/75 & 56& 0.718 & 64 & 0.821\\
OO-Any & 16/22 & 14& 0.737 & 14 & 0.737\\
\midrule
All & 97/97 & 70 & 0.722 & 78 & 0.804\\
\bottomrule
\end{tabular}
\caption{Inter-annotator agreement between the two annotators for link annotation in (a) the original URL and (b) the grouped URL settings. The ``In/OO-OSM'' and ``In/OO-Any'' rows indicate the results in the settings where (i) the database is limited to OSM or (ii) not. ``In-'' and ``OO-'' indicate scores for instances in and out of the database, respectively.} \label{tab:iaa_link}
\end{table}

As an IAA measure for link annotation, we calculated F1 score of OSM entry (or other web page) assignment for the same entities between two annotators (W1 and W2), which is similar to cluster-level hard F1 score \cite{zaporojets-etal-2022-towards}.
We evaluated link agreement only for clusters in which all members matched between two annotators' results.\footnote{As a result of adopting the editable policy of clusters (\S\ref{sec:guideline_link}), W1 and W2 merged or split clusters for three and one entities through the five documents, respectively, although the same coreference information were provided.
}
Table~\ref{tab:iaa_link} shows the F1 scores along with the numbers of annotated entities by W1, W2, and both (M).
We used two settings about identifying assigned URLs.
The first is (a) the original URL setting that  compares raw URL strings assigned by the annotators.
The second is (b) the grouped URL setting.
This treats OSM entries (web pages) representing practically the same real-world locations as the same and compares the grouped URL sets instead of original URLs.\footnote{An author manually judged the identities of different OSM entries and web pages for 34 entities unmatched between two annotators. However, the grouping process of different OSM entries can be automated based on OSM tag information.}
In addition, we used two settings about the database: (i) only OSM and (ii) any of the databases specified in \S\ref{sec:guideline_link} or other web pages.
For the 2$\times$2 settings, scores were calculated for both In-DB and OO-DB instances.\footnote{We regarded an entity as a matched In-DB instance when both annotators assigned the same URL and as a matched OO-DB instance when both annotators assigned \texttt{NOT\_FOUND} tag.}

We obtained In-DB F1 scores of 0.842 and 0.821 in the (b)-(i) grouped and In-OSM and (b)-(ii) grouped and In-Any settings, respectively.
The lower F1 scores (less numbers of matched entities) in (a) the original setting is because the annotators assigned similar but different OSM entries (web pages), which referred to practically the same locations.
For example, two annotators assigned a little bit different entries for the entity \ja{JR新宿駅} \textit{JR shinjuku eki} `JR Shinjuku Station,'\footnote{This entity consisted of a mention that occurred in a sentence: ``\ja{バスターミナルは\underline{JR新宿駅}新南口改札の下にあります。}'' (The bus terminal is located under the New South Exit of \underline{JR Shinjuku Station}.).} both of which represented Shinjuku Station operated by JR East; one entry (\href{https://www.openstreetmap.org/node/6283002002}{\texttt{node/6283002002}}) has additional tag specifying the railway line for Narita Express whereas the other entry (\href{https://www.openstreetmap.org/node/2389061844}{\texttt{node/2389061844}}) has no tags about lines.


\subsection{Summary}
We investigated IAA scores for the three annotation tasks:
(1) F1 score of 0.832 for mention annotation (all mentions),
(2) CoNLL score of 0.858 for coreference relation annotation (original clusters setting), and 
(3) In-DB F1 score of 0.842 for OSM entry assignment (grouped and In-OSM setting).
These results suggest the practical quality of our dataset in terms of the consistency.



\section{Related Work}
For more than two decades, much effort has been devoted to developing annotated corpora for English entity analysis tasks, including NER~\cite{tjong-kim-sang-2002-introduction,ling2012fine,baldwin-etal-2015-shared}, anaphora/coreference resolution~\cite{grishman-sundheim-1996-message,doddington-etal-2004-automatic,pradhan-etal-2011-conll,ghaddar-langlais-2016-wikicoref}, and ED/EL~\cite{mcnamee-etal-2010-evaluation,hoffart-etal-2011-robust,ratinov-etal-2011-local,rizzo-etal-2016-making}.
Also for Japanese, annotated corpora have been developed for general NER~\cite{sekine-etal-2002-extended,hashimoto-nakamura-2010-kakucho,iwakura-etal-2016-constructing}, (anaphoric) coreference resolution~\cite{kawahara-etal-2002-construction,hashimoto-etal-2011-construction,hangyo-etal-2014-building}, and EL~\cite{jargalsaikhan-etal-2016-building,murawaki-mori-2016-wikification}.

For English geoparsing, annotated corpora have been developed and used as benchmarks for system evaluation.
The Local Global Corpus~\cite{lieberman-etal-2010-geotagging}, TR-News~\cite{kamalloo-etal-2018}, and GeoWebNews~\cite{gritta-etal-2020-pragmatic} contains approximately 100--600 news articles from global and local news sources.
GeoVirus~\cite{gritta-etal-2018-melbourne} comprises 229 WikiNews articles focusing on viral infections.
The SemEval-2019 Task 12 dataset~\cite{weissenbacher-etal-2019-semeval} comprises 150 biomedical journal articles on the epidemiology of viruses.
GeoCorpora~\cite{wallgrun-etal-2018-geocorpora} comprises 1,639 tweet posts.
For Japanese geoparsing, \citet{matsuda-etal-2017-geographical} constructed the LRE corpus, comprising 10,000 Japanese Twitter posts, of which 793 had geo-entity-related tags.

\section{Conclusion}
This paper presented our dataset suitable for document-level geoparsing, along with the design policies, annotation guidelines, detailed dataset statistics, and inter-annotator agreement evaluation.
In the future, we plan to (1) evaluate existing systems for EL/geoparsing on our dataset and develop a document-level geoparser, and (2) enhance our dataset with additional semantic information, such as movement trajectories of travelogue writers, for more advanced analytics.
Other possible directions include (3) the construction of annotated travelogue datasets in other languages by extending our annotation guidelines.

\section*{Acknowledgments}
This study was supported by JSPS KAKENHI Grant Number JP22H03648.

\bibliographystyle{acl_natbib}

\end{document}